\documentclass{article}
\usepackage{spconf,amsmath,graphicx}

\usepackage{enumitem}
\usepackage{adjustbox}
\usepackage{booktabs} 
\usepackage{url}
\usepackage{hyperref}
\usepackage{stfloats}
\usepackage{float}
\usepackage{placeins}
\usepackage{amssymb}
\usepackage{amsmath}  
\usepackage{hanging}
\usepackage{tabto}

\setlist{nosep, leftmargin=14pt}

\usepackage{mwe} % to get dummy images
\usepackage{subcaption} % For subfigures

\title{J-CaPA : Joint Channel and Pyramid Attention \\ Improves Medical Image Segmentation}

\name{Marzia Binta Nizam\thanks{Corresponding Author: manizam@ucsc.edu} \hspace{1cm} Marian Zlateva \hspace{1cm} James Davis}
\address{Department of Computer Science, University of California, Santa Cruz, CA}

\begin{document}
%\ninept
%
\maketitle
\begin{abstract}
Medical image segmentation is crucial for diagnosis and treatment planning. Traditional CNN-based models, like U-Net, have shown promising results but struggle to capture long-range dependencies and global context. To address these limitations, we propose a transformer based architecture that jointly applies Channel Attention and Pyramid Attention mechanisms to improve multi-scale feature extraction and enhance segmentation performance for medical images. Increasing model complexity requires more training data, and we further improve model generalization with CutMix data augmentation. Our approach is evaluated on the Synapse multi-organ segmentation dataset, achieving a 6.9\% improvement in Mean Dice score and a 39.9\% improvement in Hausdorff Distance (HD95) over an implementation without our enhancements. Our proposed model demonstrates improved segmentation accuracy for complex anatomical structures, outperforming existing state of the art methods.

\end{abstract}

\section{Introduction}
\vspace{-1em} 
\label{sec:intro}
Medical image segmentation is a fundamental task in clinical applications, providing the precise identification of anatomical structures critical for various diagnoses and treatments. However, it remains challenging due to the varying size, shape, and appearance of different organs and pathologies. While convolutional neural networks (CNNs), like U-Net\cite{Unet} and its variants\cite{KiUNet,ResUnet,R2Unet}, have shown strong results, they often struggle to model long-range dependencies and global context effectively—key factors for segmenting complex structures.

To address these limitations, Transformer-based architectures have been explored in computer vision. Transformers use self-attention mechanisms to capture global context and long-range dependencies, which offers advantages over traditional CNNs. The Vision Transformer (ViT) and its adaptations have demonstrated significant improvements in medical image segmentation.

We propose a method called J-CaPA that integrates joint attention mechanisms—Channel Attention and Pyramid Attention—into a Transformer-based U-Net model. These mechanisms improve feature representation by effectively capturing both local and global contexts. These attention layers increase the number of weights, so we increase data augmentation using CutMix to insure the model does not overfit.

%Additionally, we employ CutMix data augmentation to enhance model generalization and robustness.

Our approach is evaluated on the Synapse\footnote{\url{https://www.synapse.org/Synapse:syn3193805/wiki/89480}} multi-organ segmentation dataset, which consists of 30 abdominal CT scans with annotations for several organs, including the aorta, gallbladder, kidneys, liver, pancreas, spleen, and stomach. Our model achieved a 6.9\% improvement in Mean Dice score and a 39.9\% reduction in Hausdorff Distance (HD95) over an implementation without our enhancements. Notably, the segmentation of organs such as the gallbladder, kidneys, and pancreas showed marked improvement, indicating the model's ability to handle complex anatomical structures effectively.

Evaluation results indicate that our architecture performs better than multiple state of the art comparison methods. Importantly, ablation studies show that neither Channel Attention nor Pyramid Attention used independently results in optimal performance. 
The key contribution of this paper is showing that jointly applying these attention models is what leads to enhanced medical image segmentation accuracy.
\vspace{-1.5em} 
\section{Related Work}
\vspace{-1em} 
\subsection{CNN-Based Methods for Medical Image Segmentation}
CNNs, including FCNs \cite{FCN} and U-Net variants \cite{Unet}, have shown strong segmentation performance. U-Net++ \cite{Unet++} narrows the semantic gap using dense skip connections, while Attention U-Net \cite{attentionUnet} employs attention gates for feature selection. Models like Res-UNet \cite{ResUnet} and R2U-Net \cite{R2Unet} introduced residual connections to refine segmentation tasks, while DoubleU-Net \cite{DoubleUnet} and specialized networks like PraNet \cite{pranet} and KiU-Net \cite{KiUNet} targeted specific medical challenges. Despite these advances, CNNs still struggle with modeling long-range dependencies. Attempts to integrate self-attention mechanisms \cite{wang2018non, schlemper2019attention} improved feature selection but were insufficient in capturing global context. This limitation has shifted interest towards Transformer-based architectures, which excel in combining global and local features.
\vspace{-1em} 
\subsection{Transformer-Based Approaches for Medical Image Segmentation}
\vspace{-0.5em} 
Transformers, originally developed for natural language processing \cite{vaswani2017attention}, have recently gained traction in computer vision tasks due to their ability to model long-range dependencies. Vision Transformer (ViT) \cite{ViT} demonstrated state-of-the-art performance, leading to adaptations for medical image segmentation, such as TransUNet \cite{TransUnet}, which combines the strengths of both Transformers and CNNs. Further improvements were made with models like Swin-Unet \cite{SwinUnet} and DS-TransUNet \cite{DStransUnet}, which introduced hierarchical Transformer blocks and enhanced feature fusion. Other notable models, such as AA-TransUNet \cite{AAtransUnet} and DA-TransUNet \cite{DATransUnet}, explored spatial and channel attention mechanisms.

\begin{figure*}[ht]
  \centering
  \includegraphics[width=0.9\linewidth]{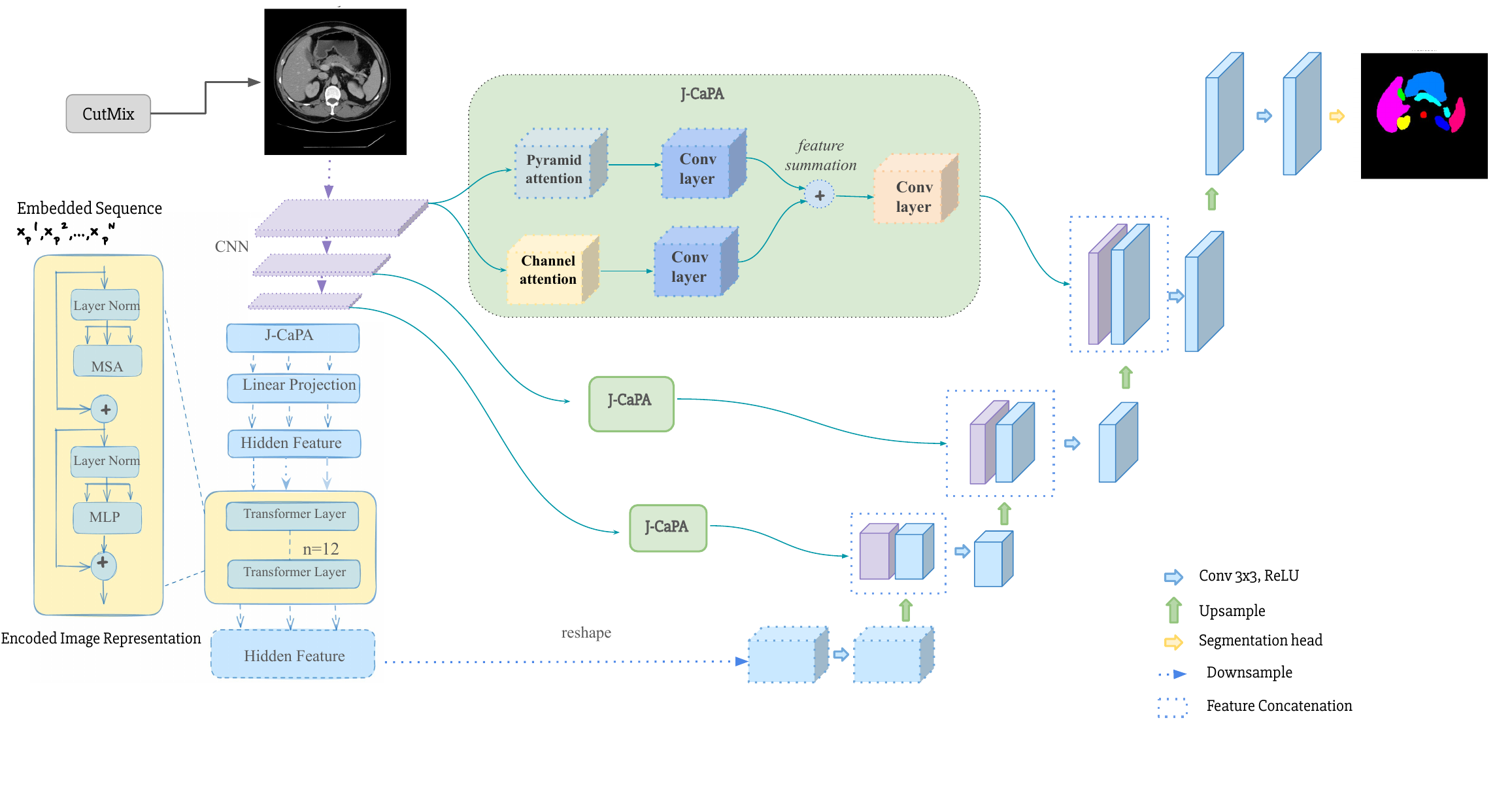}
  \vspace{-0.2cm}
  \caption{Overview of the proposed framework. Input medical images are processed by a CNN-based encoder, followed by Transformer layers and our Joint Attention blocks (combining Pyramid and Channel Attention). Features at multiple scales are refined by Joint Attention and passed through skip connections to the decoder. The decoder performs CNN-based up-sampling to generate high-resolution segmentation maps, capturing detailed anatomical structures.}
  \label{fig:methodology}
\end{figure*}

Transformer base large models, including Segment Anything Model (SAM) \cite{SAM}, SegGPT \cite{wang2023seggpt}, and STU-Net \cite{STU-net}, have gained attention for their zero-shot generalization capabilities. However, their performance in medical image segmentation has been limited due to a lack of domain-specific training data. Adaptations like MedSAM \cite{MedSAM} and Medical SAM Adapter \cite{zhang2024segment} have fine-tuned SAM for medical datasets through prompt-based methods, while prompt-free approaches such as AutoSAM \cite{AutoSAM} have attempted to eliminate the need for prompts. SAMed \cite{SAMed} enhances SAM's segmentation performance by integrating LoRA layers \cite{hu2021lora} into the model. 

Our approach builds on these advancements by integrating Channel and Pyramid Attention into a transformer-based architecture, enabling multi-scale feature extraction tailored for medical image segmentation.

%Unfortunately medical applications often require training data transparency to provide model trust. Large vision models by definition contain a large corpus of unknown and general purpose inputs, making the relation between input data and outputs much more difficult to understand. 
\vspace{-1em} 
\section{Methodology}
\vspace{-1em} 
\begin{table*}[!htbp]
\centering
\caption{Segmentation accuracy of different methods on the Synapse multi-organ CT
dataset(Average Dice Similarity Coefficient (DSC) score (\%) and average Hausdorff Distance (HD) in mm, along with DSC score (\%) for each organ). The best results are highlighted in bold.}
\label{tab:model_performance}
%\scriptsize
\LARGE
\setlength{\tabcolsep}{8pt} % Adjust column separation
\renewcommand{\arraystretch}{1.2} % Adjust row height
\resizebox{\textwidth}{!}{%
\begin{tabular}{lcccccccccc} 
\toprule
\textbf{Model} & \textbf{Mean Dice $\uparrow$} & \textbf{Mean HD95 $\downarrow$} & \textbf{Aorta} & \textbf{Gallbladder} & \textbf{Kidney (L)} & \textbf{Kidney (R)} & \textbf{Liver} & \textbf{Pancreas} & \textbf{Spleen} & \textbf{Stomach} \\
\midrule
TransUNet \cite{TransUnet} & 76.90 & 32.87 & 86.80 & 59.60 & 81.40 & 74.00 & 94.50 & 54.10 & 87.30 & 77.80 \\
TransNorm\cite{transnorm} & 78.40 & 30.25 & 86.23 & 65.10 & 82.18 & 78.63 & 94.22 & 55.34 & 89.50 & 76.01 \\
SwinUnet \cite{SwinUnet} & 79.13 & 21.55 & 85.47 & 66.53 & 83.28 & 79.61 & 94.29 & 56.58 & 90.66 & 76.60 \\
DA-TransUnet \cite{DATransUnet} & 79.80 & 23.48 & 86.54 & 65.27 & 81.70 & 80.45 & 94.57 & 61.62 & 88.53 & 79.73 \\
IB-TransUNet\cite{IBTransUnet} & 81.05 & 22.63 & 88.24 & 66.25 & 83.89 & 79.87 & 94.63 & 63.56 & 90.23 & 81.75 \\
SAMed\cite{SAMed} & 81.88 &	20.64 &	87.77 &	\textbf{69.11}	& 80.45	& 79.95 &	\textbf{94.8} & \textbf{72.17} & 88.72 & 82.06 \\
\textbf{Ours} & \textbf{82.29} & \textbf{19.74} & \textbf{88.28} & 63.10 & \textbf{86.00} & \textbf{83.10} & 94.60 & 69.30 & \textbf{90.70} & \textbf{83.30} \\
\bottomrule
\end{tabular}%
}
\end{table*}
\subsection{Model Architecture Overview}
\vspace{-0.5em} 
The overall architecture of our model is a Transformer based structure, with an encoder-decoder design as 
shown in Figure~\ref{fig:methodology}. The encoder employs Transformer blocks to capture global context, while the decoder reconstructs detailed segmentation maps. 
%The model uses an encoder-decoder structure similar to the TransUNet, incorporating Transformer-based modules for global context modeling. 
The encoder features a modified ResNetV2 backbone that extracts hierarchical features through convolutional layers and residual bottleneck blocks. These features are refined by joint attention modules to enhance both local and global context. The decoder then uses upsampling operations to reconstruct segmentation maps from the refined features provided by the attention modules.
\vspace{-1em} 
\subsection{Joint Attention Mechanisms}
\vspace{-0.5em} 
To enhance feature representation, we incorporate two types of attention mechanisms, Channel Attention and Pyramid Attention. 
\vspace{-1em} 
\subsubsection{Channel Attention Module}
\vspace{-0.5em} 
The Channel Attention Module (CAM) was originally introduced for segmenting city street scenes~\cite{fu2019dual} and has also been successfully applied in medical image segmentation tasks~\cite{li2021ta,DATransUnet}. For our work, we take an input feature map \( X \in \mathbb{R}^{B \times C \times H \times W} \), where B is the batch size, C is the number of channels, and H and W are the height and width of the feature map, respectively. CAM computes inter-channel dependencies by reshaping this feature map into \( B \times C \times (H \times W) \). For each channel, query and key representations are derived by projecting the input into a pair of matrices. Following previous work~\cite{DATransUnet}, the energy matrix is normalized using a max-value subtraction technique, and softmax is applied to obtain attention weights. These weights are then used to reweight the value representation of the input, which is then reshaped backed to \( B \times C \times H \times W \). The final output is obtained by blending the attention-modified features with the original input, controlled by learnable parameter \( \gamma_{\text{CA}} \).  Initially set to zero,  \( \gamma_{\text{CA}} \) is updated during training through backpropagation. As the model trains, \( \gamma_{\text{CA}} \) is optimized using gradient descent.
\vspace{-1.5em} 
\subsubsection{Pyramid Attention Module}
\vspace{-0.5em} 
The Pyramid Attention Module captures multi-scale context by applying attention across different spatial scales~\cite{yu2021multi}. The module processes the input feature map \( X \in \mathbb{R}^{B \times C \times H \times W} \), denoted as \(X_s\),  where \( s \) represents the scale factor (original size: s=1, half-size: s=0.5, and quarter-size: s=0.25). For each scale \( s \), we apply three convolutional layers to produce the \textit{query}, \textit{key}, and \textit{value} representations:
\[
Q_s = W_Q X_s, \quad K_s = W_K X_s, \quad V_s = W_V X_s
\]
where \( W_Q, W_K \in \mathbb{R}^{C \times C/8} \) are learnable weight matrices, and \( W_V \in \mathbb{R}^{C \times C} \) preserves the original dimensions. 
Attention weights are computed using dot-product attention, followed by softmax normalization, to capture contextual relationships between spatial positions at each scale. The attention-modified features are then upsampled back to the original resolution and combined with the input feature map using a learnable parameter, \( \gamma_{\text{PA}} \). Initially set to zero,  \( \gamma_{\text{PA}} \) is updated during training through backpropagation, similar to the Channel Attention Module, but optimized independently for the Pyramid Attention Module.
\vspace{-1em} 
\subsection{Feature Fusion and Reconstruction}
\vspace{-0.5em} 
%The decoder combines the outputs from the Pyramid and Channel Attention modules using convolutional layers, refining the feature maps. These refined features are further processed through additional convolutions and upsampling layers to reconstruct the segmentation masks. The final layer, activated by ReLU, generates the segmentation predictions.%
Within the J-CaPA module, the Pyramid Attention (PA) and Channel Attention (CA) process the input feature map independently. The output of PA, \( F_{\text{PA}} \in \mathbb{R}^{B \times C \times H \times W} \), and output of CA, \( F_{\text{CA}} \in \mathbb{R}^{B \times C \times H \times W} \), are fused via element-wise summation:
\[
F_{\text{fused}} = F_{\text{PA}} + F_{\text{CA}}
\]
The fused map is then passed through 3x3 convolutional layers to refine the features and capture more details. After the convolutional layers, upsampling layers using bilinear interpolation are applied to restore the feature map to the original input resolution.  Finally, a ReLU-activated convolutional layer generates the segmentation mask. 
\vspace{-1.5em} 
\subsection{Extended Data Augmentation}
\vspace{-1em} 
Using two attention mechanisms in our model increases the size of the model, requiring an increased amount of training data to support generalization and avoid over-fitting. Thus we extend data augmentation to include CutMix augmentation which combines patches from different images~\cite{yun2019cutmix}. Segments from one image are randomly cut and pasted onto another, with their respective labels. The size of the segments is varied randomly, covering between \(20\%\) to \(60\%\) of the image area, ensuring diverse learning scenarios. CutMix was applied to \(33\%\) of the images in each training batch to maintain a balance between original and augmented data. For the remaining images in each batch, we applied standard augmentation techniques, such as flipping and rotating.
\vspace{-1em} 
%\section{Experiments}
\subsection{Data}
\vspace{-0.5em} 
Experiments in our study use the Synapse multi-organ segmentation dataset, which contains 30 abdominal CT scans. These scans consist of a total of 3,779 axial contrast-enhanced clinical CT images, where each CT scan is composed of multiple slices. The dataset provides annotations for several organs across these scans, including the aorta, gallbladder, left kidney, right kidney, liver, pancreas, spleen, and stomach, along with a ‘none’ label, resulting in a total of nine target classes. We used a preprocessed version of this dataset~{\cite{TransUnet}. We follow prior work to split the 30 scans, with 18 used for training, and the remaining 12  reserved for testing. 
\vspace{-1em} 
\subsection{Experimental Details}
\vspace{-0.5em} 
We use a baseline transformer which prior work indicated was the most effective among configurations tested~\cite{TransUnet}. The specific configuration is called R50-ViT-B\_16.
In order to allow a fair comparison across methods, all parameters, including model architecture and training settings, were kept consistent with original specifications. The model was trained for 150 epochs with a batch size of 8 on a single RTX 3080 GPU, taking approximately 2 hours to complete the training.
\vspace{-1em}
\section{Results}
\vspace{-0.5em} 
%Upon successfully replicating the R50-ViT-B\_16 configuration of the TransUnet model, we proceeded to evaluate its performance using the standard metrics from the original study.  Upon integrating the dual attention block and the CutMix augmentation into the TransUnet model, 
We conducted an evaluation to compare the performance of our model to multiple state-of-the-art methods when segmenting organs in abdominal CT scans. Results are provided in Table~\ref{tab:model_performance}. We follow prior work and report results using the metrics Mean Dice score and Hausdorff Distance (HD95), facilitating a direct comparison with baseline models. Our method outperforms all comparison models on these metrics.

\begin{figure}[b!] 
\setlength{\textfloatsep}{10pt} 
\centering
\resizebox{1.0\linewidth}{!}{%
    \includegraphics{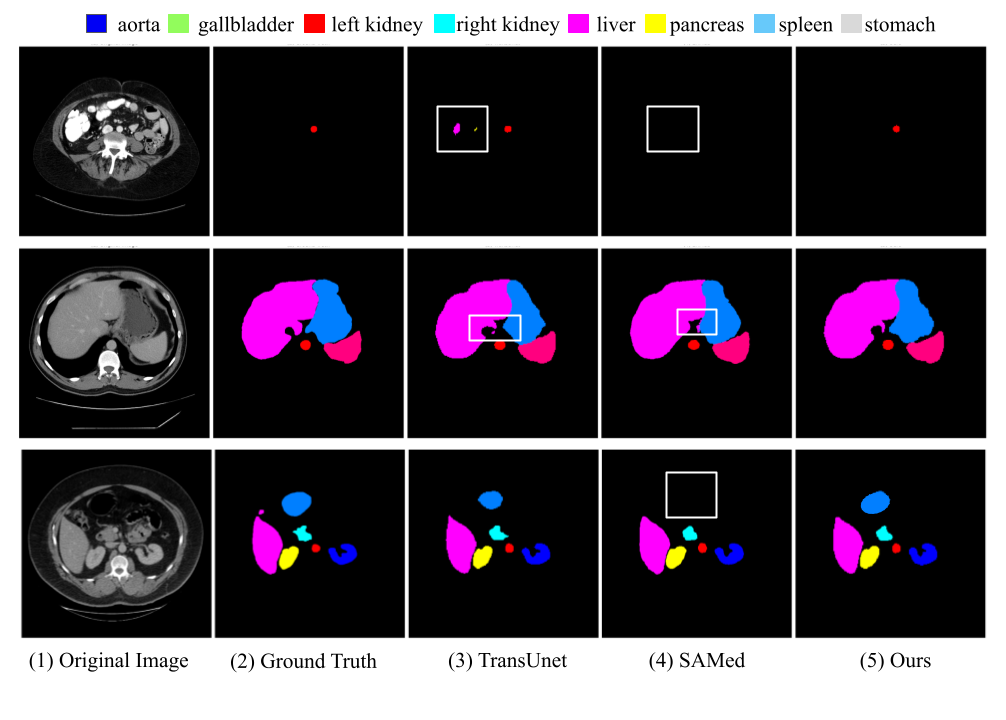}
    
}

\caption{Visual comparison of segmentation results on the Synapse dataset. Each row represents a different case, with the columns showing: (1) the original CT image, (2) ground truth segmentation labels, (3) predictions from Trans\-UNet, and (4) predictions from SAMed (5) predictions from our proposed model. }
\label{fig:viz}
\end{figure}

The integration of Joint Channel and Pyramid Attention mechanisms into our architecture significantly enhanced our model's segmentation capabilities. %Baseline performance without these enhancements is equivalent to the row labeled TransUNet.%
A higher Mean Dice score reflects better overlap between predicted and ground truth segmentations, while a lower Hausdorff Distance (HD95) indicates more precise organ boundary delineation. Our method achieved a 6.9\%\ improvement in Mean Dice, increasing it to 82.29\%\, and a \(39.9\%\) reduction in HD95, lowering it to 19.743 mm, compared to an implementation without the channel and pyramid enhancements, identical to TransUnet in this table.

%In general, we expect access to larger training sets to improve model performance. All of the UNet based comparison methods have access to the same underlying training data set as our model, and are thus a fair comparison of the methods themselves. SAMed is based on the large vision model SAM, and thus trained on a larger dataset with more than a billion annotations. %

Figure \ref{fig:viz} presents a visual comparison of segmentation performance across cases from the Synapse dataset. The white boxes highlight areas where TransUNet and SAMed results contain false positives or false negatives, failing to capture certain structures accurately. In contrast, our model demonstrates improved precision by successfully identifying these challenging regions. For instance, in the first row, Trans\-UNet produced a false positive in the liver, while SAMed failed to predict the left kidney. In the last image, SAMed missed the spleen entirely, underscoring our model’s ability to correct these misclassifications.

\vspace{-1em} 
\section{Ablation Study}
\vspace{-0.5em} 
We conducted an ablation study to assess the impact of Channel Attention, Pyramid Attention, and CutMix data augmentation on our model’s performance. The results, shown in Table~\ref{tab:ablation_study}, compare a baseline implementation without our enhancements to the model with each enhancement added. We find that while each of Channel Attention and Pyramid Attention improve on the baseline, neither acting alone performs as well as the Joint Attention model. 

Our full model, combining Joint Channel and Pyramid Attention and enhanced data augmentation using CutMix, achieves the best overall performance with a Mean Dice of 82.2\%\ and HD95 of 19.74 mm.
 
\setlength{\tabcolsep}{5pt} % Reduce column separation to save space
\renewcommand{\arraystretch}{1.2} % Adjust row height
\begin{table}[ht]
\centering
\caption{Ablation study results for Mean Dice score and Hausdorff Distance (HD95).}
\label{tab:ablation_study}
\small % Adjust font size to make it smaller if needed
\begin{tabular}{p{4.2cm}p{2cm}p{2cm}} 
\toprule
\textbf{Model} & \textbf{Mean Dice} & \textbf{Mean HD95} \\
\midrule
Baseline (without enhancements) & 76.90 & 32.87 \\
Baseline +Channel Attention & 79.70 & 23.56 \\
Baseline +Pyramid Attention & 78.14 & 31.38 \\
Baseline +Joint Attention & 80.30 & 23.73 \\
Baseline +CutMix & 79.80 & 22.69 \\
\textbf{Ours (Full Model)} & \textbf{82.29} & \textbf{19.74} \\
\bottomrule
\end{tabular}
\end{table}

\vspace{-1em} 
\section{Conclusion}
\vspace{-0.5em} 
% can we merge with other methods for even better performance?
% data aug alone doesnt do it, but will even more data augmentation help
In this paper, we propose a Transformer-based U-Net model for medical image segmentation, integrating Joint Channel and Pyramid Attention mechanisms (J-CaPA) to improve feature representation.  Our approach demonstrates notable improvements in segmentation accuracy and generalization, particularly for challenging organs in abdominal CT scans. 
\vspace{-1em} 
\section{Compliance with ethical standards}
\vspace{-1em} 
The Synapse multi-organ segmentation dataset used in this study is publicly available and contains de-identified, anonymized data. No further ethical approval was required. 
\vspace{-1em} 
\section{Acknowledgement}
\vspace{-1em} 
We are grateful to Prof. Yuyin Zhou for her guidance and suggestions throughout the duration of this project, and Vanshika Vats for her valuable feedback. No financial conflicts exist.
\vspace{-1em}

%% ---------------------------------------------------------

%%
%% This is the original and correct bibliography
%% ---------------------------------------------------------
%\begingroup
\small% Shrinks the font size for the references
%\footnotesize
%\setlength{\itemsep}{0pt} % Removes extra space between references

\bibliographystyle{IEEEbib}
\bibliography{refs}
%\endgroup
%% ---------------------------------------------------------

\end{document}